\documentclass[twocolumn]{article}

\usepackage[T1]{fontenc}
\usepackage[utf8]{inputenc}
\usepackage{textcomp}
\usepackage{graphicx}
\usepackage{url}
\usepackage{amsmath}
\usepackage{amssymb}
\usepackage{amsfonts}
\usepackage{xcolor}
\usepackage{array}
\usepackage{float}
\usepackage{subcaption}
\usepackage{caption}
\usepackage{booktabs}
\usepackage{geometry}
\usepackage{multirow}
\usepackage{placeins}
\usepackage{hyperref}
\usepackage{authblk}
\usepackage{enumitem}
\usepackage{algorithm}
\usepackage{algpseudocode}
\usepackage{tikz}
\usetikzlibrary{shapes.geometric,arrows.meta,positioning,fit,backgrounds,calc}

\title{3D Cell Morphology--Clinical Text Contrastive Learning
       with Voxel Swin Transformer and Finding Encoder Domain-Adaptation}

\author[1]{Takato Yasuno}
\date{}

\begin{document}

\renewcommand{\abstractname}{Abstract}
\renewcommand{\refname}{References}
\renewcommand{\figurename}{Figure}
\renewcommand{\tablename}{Table}

\maketitle

\begin{abstract}

Our prior MVP (v0.2.0) adapted a 2D CLIP dual encoder with LoRA to a
multi-source cell-microscopy corpus, reaching anomaly-scoring ROC-AUC up
to 0.985. This paper advances the same contrastive-learning
domain-adaptation methodology from 2D microscope \emph{images} to 3D
microscope \emph{volumes}, pairing a 3D Swin Transformer voxel encoder
(MONAI's \texttt{SwinUNETR} backbone) with
ClinicalBERT~\cite{alsentzer2019publicly} as the finding-text encoder. We
train on 3{,}000 single-cell 3D confocal fluorescence volumes (4 tagged
organelle structures) from the Allen Institute's public hiPSC single-cell
image dataset~\cite{viana2023integrated, alleninstitute2021hipsc}, paired
with template-generated finding text, with mitotic (dividing) cells
oversampled as an abnormal-like proxy class validated by ROC-AUC. Naively
reusing v0.1.0/v0.2.0's ``freeze backbone, LoRA the attention
projections'' recipe for the Voxel Encoder collapses voxel embeddings to
near-input-invariance, since LoRA's low-rank delta cannot compensate for
an otherwise-frozen, \emph{randomly initialized} (not pretrained) 3D Swin
backbone, even as InfoNCE loss decreases healthily. Switching the Voxel
Encoder to full fine-tuning resolves this: 4-way zero-shot structure
accuracy rises from 25.0\% (chance) to 48.6\%, and voxel anomaly-scoring
ROC-AUC improves from 0.486 to a peak of 0.628 (still below the 0.70
sanity threshold), while bidirectional retrieval recall stays low
(R@10 $\le$3.8\%). We further introduce a nearest-neighbor
``diagnosis-reason retrieval'' evaluation, reaching 29.8\% hit-rate@\{1,5\}
at its best scale while scoring 0.0\% elsewhere. The central finding of
this MVP is architectural: LoRA's parameter-efficiency assumption
presumes a pretrained backbone worth preserving, and breaks down when the
backbone is randomly initialized. A supplementary extension (v0.3.1,
reported in the Supplementary Material) tests whether this fix
generalizes to greater structural diversity and scale (4$\to$8
structures, 3,000$\to$8,000 cells) and finds that it does not: the same
voxel-embedding collapse recurs despite an even lower training loss,
showing that training-loss health alone does not certify that a
contrastive-learning fix generalizes to a new diversity/scale regime.

\end{abstract}


\section{Introduction}
\label{sec:introduction}

Our MVP progression develops a contrastive-learning domain-adaptation
methodology for material damage understanding using proxy data where
production road-pavement microscopy is not yet available. v0.1.0~\cite{yasuno2026materialvlm}
established a 2D CLIP dual encoder (LoRA, symmetric InfoNCE) on
macro-texture photographs (DTD) as a proxy; v0.2.0 moved to genuine 2D
biological cell micrographs (NIH Malaria, C-NMC 2019, LDC), reusing the
same architecture and reaching cell-anomaly-scoring ROC-AUC up to 0.985
against real within-domain ground truth. Both, however, remained
strictly \emph{2D}, with CLIP's native 2D vision tower as the encoder --
whereas the eventual production task, diagnosing durability-relevant
anomalies from 3D microscopy (e.g., X-ray CT or confocal 3D surface
scans), is inherently volumetric, and its accompanying diagnostic
``finding'' is naturally expressed as free text rather than a short
class label.

This paper (v0.3.0) closes both gaps simultaneously. Because no public
3D microscopy dataset of pavement-material damage yet exists, we
substitute the Allen Institute's public hiPSC single-cell 3D confocal
fluorescence image dataset~\cite{viana2023integrated,
alleninstitute2021hipsc} as a 3D-voxel, multi-organelle biological
proxy, and replace both towers of the v0.1.0/v0.2.0 CLIP architecture: a
3D Swin Transformer~\cite{liu2021swin,hatamizadeh2022swin} voxel encoder
in place of the 2D CLIP vision tower, and
ClinicalBERT~\cite{alsentzer2019publicly} -- a BERT-base model
pretrained on real MIMIC-III clinical notes -- in place of the CLIP text
tower, yielding \texttt{VoxelSwinCLAIP}.

Our contributions are:
\begin{itemize}[leftmargin=*,itemsep=2pt]
  \item \textbf{\texttt{VoxelSwinCLAIP}}, the first 3D extension of this
        MVP progression's dual-encoder architecture, pairing a fully
        fine-tuned 3D Swin Transformer Voxel Encoder with a
        LoRA-adapted ClinicalBERT Finding Encoder under the same
        symmetric InfoNCE objective, together with an
        empirically-discovered adaptation-strategy correction (full
        fine-tuning, not LoRA, for the non-pretrained Voxel Encoder;
        Section~\ref{sec:lessons2}).
  \item A new \textbf{3D-voxel proxy dataset} built from the Allen
        Institute's hiPSC single-cell image collection, spanning four
        organelle/structure categories (nuclear envelope, mitochondria,
        microtubules, tight junctions), with mitotic (dividing) cells
        deliberately oversampled as the abnormal-like proxy class.
  \item A new \textbf{diagnosis-reason retrieval} evaluation that
        retrieves an explanatory finding text for an anomalous volume,
        operationalizing this MVP series' long-term ``Medical CLAIP''
        goal (Figure~\ref{fig:architecture}) within a retrieval-only,
        non-generative MVP scope.
  \item A set of \textbf{data-engineering corrections} -- a
        channel-index metadata scope mismatch and a
        candidate-structure-name mismatch (``CAAX''/``TOM20'') -- both
        caught only by inspecting the real Allen Institute metadata at
        implementation time, documented as a cautionary methodological
        lesson for reusing biomedical imaging metadata without direct
        verification (Section~\ref{sec:lessons}).
\end{itemize}

\section{Related Work}
\label{sec:related}

\subsection{3D Vision Transformers for Volumetric Medical Imaging}

Swin Transformer~\cite{liu2021swin} introduced hierarchical, shifted-window
self-attention for 2D vision; Swin UNETR~\cite{hatamizadeh2022swin}
extended this to 3D volumetric medical segmentation (brain tumor MRI),
pairing a hierarchical 3D Swin encoder with a convolutional UNETR-style
decoder. We reuse only the encoder half of MONAI's~\cite{cardoso2022monai}
\texttt{SwinUNETR} implementation (\texttt{swinViT}), discarding the
segmentation decoder entirely, since our task is contrastive representation
learning rather than per-voxel segmentation (Section~\ref{sec:voxeltower}).
Several other works adapt Transformers to volumetric medical imaging:
UNETR~\cite{hatamizadeh2022unetr} first reformulated 3D segmentation as a
sequence-to-sequence prediction problem with a plain ViT encoder (no
window hierarchy); TransBTS~\cite{wang2021transbts} instead hybridizes a
3D CNN encoder with a Transformer bottleneck for brain-tumor MRI
segmentation; nnFormer~\cite{zhou2023nnformer} interleaves convolution
with local/global volume-based self-attention within a U-Net-like
skip-attention design; and 3D UX-Net~\cite{lee2023ux3dnet} replaces
Swin's window self-attention with large-kernel depth-wise convolutions to
approximate its receptive field at lower parameter cost. The
shifted-window mechanism we inherit via Swin UNETR's \texttt{swinViT}
further traces to Video Swin Transformer~\cite{liu2022videoswin}, which
extended 2D Swin's local window attention to a third (temporal) axis for
video recognition -- the same 3D-windowing principle \texttt{SwinUNETR}
instead applies to the depth axis of volumetric medical images.

\subsection{Clinical Language Models for Vision-Language Alignment}

ClinicalBERT~\cite{alsentzer2019publicly} is a BERT-base model continued
from pretraining on real MIMIC-III intensive-care clinical notes,
widely used as a text encoder in medical vision-language and
report-generation systems. We use it here as our Finding Encoder even
though -- unlike a genuine medical CLIP/CLAIP deployment -- our
biological-proxy ``finding text'' is template-generated rather than a
real inspection report (Section~\ref{sec:dataset}); ClinicalBERT's
clinical-language prior is intended to ease a future transition to real
pavement-inspection reports once they become available, consistent with
this MVP progression's proxy-data philosophy. ClinicalBERT belongs to a
broader family of domain-adapted BERT variants: BioBERT~\cite{lee2020biobert}
continues pretraining from general-domain BERT on PubMed abstracts and
PMC full texts, while PubMedBERT~\cite{gu2021pubmedbert} instead
pretrains from scratch on biomedical text, showing this outperforms
continual pretraining when the in-domain corpus is sufficiently large.
For vision-language alignment specifically,
ConVIRT~\cite{zhang2022convirt} first demonstrated bidirectional
image-text contrastive pretraining on paired chest X-rays and radiology
reports, later extended by GLoRIA~\cite{huang2021gloria} to a
global-local contrastive objective that additionally aligns image
sub-regions with report words, and by BioViL~\cite{boecking2022biovil},
which pairs a radiology-specific text encoder (CXR-BERT) with
masked-language-modelling regularization to improve textual semantic
grounding in self-supervised vision-language pretraining. We adopt the
simpler, off-the-shelf ClinicalBERT -- rather than a radiology-report-
pretrained encoder such as CXR-BERT -- because our template-generated
finding text is not yet a real radiology report (Section~\ref{sec:dataset}).

\subsection{CLIP-Based Zero/Few-Shot Anomaly Scoring}

As in v0.1.0/v0.2.0, our anomaly-scoring evaluation follows the WinCLIP
compositional prompt-ensemble pattern~\cite{jeong2023winclip}, as extended
by AnomalyCLIP~\cite{zhou2024anomalyclip}, AnomalyGPT~\cite{gu2024anomalygpt},
and related open-vocabulary and reasoning-augmented
variants~\cite{cao2023gpt4vanomaly,xu2025anomalyov,zhang2025trainingfree,
jin2025logicad}, here applied to voxel embeddings rather than 2D image
embeddings, with ``interphase'' vs.\ ``mitotic'' compositional prompts
substituting for v0.1.0's ``intact'' vs.\ ``damaged''
(Section~\ref{sec:scoring}). Recent surveys situate this
compositional-prompt, CLIP-based line of work within the broader visual
anomaly detection literature: Cao et al.~\cite{cao2024advsurvey} organize
visual anomaly detection approaches by task setting (zero-/few-shot,
weakly supervised, fully supervised) and highlight vision-language
foundation models as a rapidly growing zero-shot paradigm; Liu et
al.~\cite{liu2024iadsurvey} review industrial image anomaly detection by
network architecture and supervision level, situating WinCLIP-style
prompt ensembles alongside reconstruction- and embedding-based methods;
and Mou et al.~\cite{mou2026mmadsurvey} survey multi-modal anomaly
detection specifically, formalizing how foundation-model pretraining and
cross-modal alignment -- the same ingredients underlying our voxel/text
VoxelSwinCLAIP architecture -- are reshaping the field beyond single-image
settings.

\subsection{The Allen Institute hiPSC Single-Cell 3D Image Dataset}

Viana et al.~\cite{viana2023integrated} introduced the WTC-11 hiPSC
single-cell 3D confocal fluorescence image dataset, imaging $\sim$25
fluorescently tagged sub-cellular structures across $>$200{,}000
segmented single cells, each annotated with a combined-stage
\texttt{cell\_stage} label (\texttt{M0} = interphase;
\texttt{M1M2}/\texttt{M3}/\texttt{M4M5}/\texttt{M6M7\_single}/
\texttt{M6M7\_complete} = successive mitotic sub-stage buckets).

Among public subcellular-imaging resources, it occupies a distinctive
niche: the Human Protein Atlas Cell Atlas~\cite{thul2017subcellular} is
comparably large-scale ($>$12{,}000 proteins) but only 2D and without
cell-cycle labels, while OpenOrganelle~\cite{heinrich2021openorganelle}
is genuinely volumetric (FIB-SEM) but limited to a handful of whole-cell
volumes by EM imaging throughput. The Allen Institute dataset is, to our
knowledge, the only public resource combining 3D volumes, population-scale
single-cell counts, and live mitotic-stage labels -- a combination that
makes it well-suited as a 3D-voxel, multi-organelle biological proxy for
contrastive vision-language pretraining, a use not previously explored,
and one that closely matches this MVP series' originally envisioned
``cell organelle-level abnormal morphology'' learning target.

\subsection{Positioning of This Work}

This paper is the third MVP in a progression from supervised
text-generation fine-tuning~\cite{yasuno2026damagevlm}, through 2D
macro-texture contrastive domain adaptation~\cite{yasuno2026materialvlm}
(v0.1.0) and 2D multi-source biological-microscopy contrastive domain
adaptation (v0.2.0), to the present 3D-voxel,
clinical-text contrastive domain adaptation (v0.3.0) -- each step moving
the proxy data and architecture closer, in imaging dimensionality and
text-encoder realism, to the eventual production task of diagnosing
3D material anomalies in road-pavement microscopy via retrievable
textual explanations.

\section{Methodology}
\label{sec:methodology}

\subsection{System Architecture: VoxelSwinCLAIP}
\label{sec:voxeltower}

\begin{figure}[t]
\centering
\resizebox{\linewidth}{!}{%
\begin{tikzpicture}[
  node distance=0.6cm and 0.4cm,
  box/.style={rectangle, draw=gray!70, rounded corners=3pt,
              minimum width=2.8cm, minimum height=0.9cm,
              align=center, font=\footnotesize},
  stage/.style={box, fill=blue!12, draw=blue!50},
  proc/.style={box,  fill=orange!12, draw=orange!50},
  outbox/.style={box, fill=green!12, draw=green!50},
  arr/.style={->, >=stealth, thick,
}
]
\node[stage] (vox) {Voxel Encoder\\(3D Swin Transformer, full fine-tune)};
\node[stage, right=of vox] (txt) {Finding Encoder\\(ClinicalBERT + LoRA)};
\node[proc, below=of $(vox)!0.5!(txt)$] (sim) {Contrastive similarity\\matrix (InfoNCE)};
\node[outbox, below=of sim] (mat) {VoxelSwinCLAIP\\(domain-adapted)};
\draw[arr] (vox) -- (sim);
\draw[arr] (txt) -- (sim);
\draw[arr] (sim) -- (mat);
\end{tikzpicture}%
}
\caption{VoxelSwinCLAIP architecture: a 3D Swin Transformer voxel tower
         replaces v0.1.0/v0.2.0's 2D CLIP vision tower, and ClinicalBERT
         replaces CLIP's native text tower, both trained under the
         same symmetric InfoNCE objective (Section~\ref{sec:voxeltower}).
         The Voxel Encoder is fully fine-tuned (no pretrained prior to
         preserve); only the Finding Encoder uses LoRA, since
         ClinicalBERT is genuinely pretrained (Section~\ref{sec:lessons2}).
         Gene/structure-identity encoding (a third tower, toward a full
         Medical CLAIP) is deferred to v0.4.0 (Section~\ref{sec:limitations}).}
\label{fig:architecture}
\end{figure}
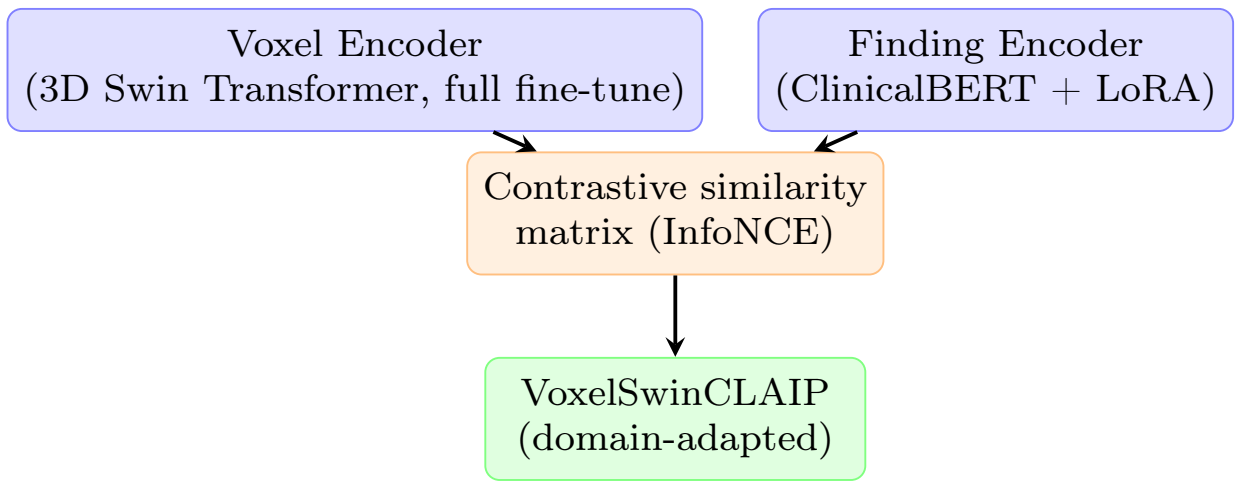

Figure~\ref{fig:architecture} shows the v0.3.0 architecture. A
\textbf{Voxel Encoder} (3D Swin Transformer, MONAI's \texttt{SwinUNETR}
encoder-only backbone \texttt{swinViT}, fully fine-tuned) and a
\textbf{Finding Encoder} (ClinicalBERT, LoRA-adapted) each produce a
512-dimensional, L2-normalized embedding; a learnable temperature
(\texttt{logit\_scale}, initialized to CLIP's $\log(1/0.07)$) scales their
dot product into a contrastive similarity matrix trained with the same
symmetric InfoNCE objective as v0.1.0/v0.2.0, yielding
\texttt{VoxelSwinCLAIP}, used for the same three downstream capabilities
as before (zero-shot classification, bidirectional retrieval,
compositional-prompt anomaly scoring) plus a new fourth capability
introduced in this version (diagnosis-reason retrieval,
Section~\ref{sec:evaluation}).

Concretely, the Voxel Encoder takes a 2-channel 64$\times$64$\times$64
volume ([DNA, structure] channels; Section~\ref{sec:dataset}) and
produces a hierarchical sequence of 5 feature maps via \texttt{swinViT};
we global-average-pool the deepest stage's spatial dimensions and project
with a linear head to the shared 512-d embedding space. Unlike the
Finding Encoder, the Voxel Encoder is \emph{fully fine-tuned} (all
parameters trainable) rather than LoRA-only-adapted: an initial LoRA-only
design (rank $r=8$, $\alpha$=16 on only each window-attention block's
\texttt{qkv}/\texttt{proj} linear layers, with \texttt{patch\_embed} and
\texttt{PatchMerging} frozen at their random initialization) produced
near-input-invariant voxel embeddings (pairwise cosine similarity
0.93--0.99 across distinct structures after a full 15-epoch training run),
because a small low-rank delta cannot carve a useful representation out of
an otherwise-frozen \emph{random} backbone -- LoRA presumes a pretrained
prior worth preserving, which the randomly initialized 3D Swin Transformer
does not have. Switching to full fine-tuning (3.68M trainable parameters
for feature\_size=24, still comfortably within the 16GB VRAM budget)
resolved this collapse, as confirmed by re-inspecting pairwise voxel-
embedding cosine similarities after retraining
(Section~\ref{sec:lessons2}). The Finding Encoder tokenizes the finding
text with ClinicalBERT's own tokenizer, LoRA-adapts (rank $r=16$,
$\alpha$=32, dropout 0.05) the \texttt{query}/\texttt{key}/\texttt{value}
projections of every self-attention layer -- appropriate here because
ClinicalBERT carries a genuine pretrained clinical-language prior worth
preserving -- and projects the \texttt{[CLS]} token's final hidden state
to the same 512-d space.

\subsection{Mathematical Formulation: From Input Volume/Text to Downstream Output}
\label{sec:mathformulation}

We now state the full pipeline of Section~\ref{sec:voxeltower} formally,
end to end: from a raw preprocessed voxel/text pair, through the Voxel
and Finding Encoders, to the training objective and the closed-form
decision rules that produce every reported downstream metric.

\paragraph{Input.}
Let $x \in \mathbb{R}^{2\times64\times64\times64}$ denote a single
preprocessed per-cell input volume (channels: DNA, structure channel;
percentile-clipped, min-max normalized to $[0,1]$, and resized to
$64^3$ as described in Section~\ref{sec:dataset}), and let
$t=(w_1,\dots,w_L)$ denote its paired finding-text token sequence,
produced by ClinicalBERT's WordPiece tokenizer from the
template-generated finding text (Section~\ref{sec:dataset}). Each
training batch consists of $N=12$ such $(x_i,t_i)$ pairs
(Table~\ref{tab:hparams}).

\paragraph{Voxel Encoder $f_V$.}
The 3D Swin Transformer backbone (feature size $C=24$) maps $x$ through
a patch-embedding stem followed by four hierarchical window-attention
stages $\ell=1,\dots,4$, each halving spatial resolution and doubling
channel width:
\begin{equation}
  z_0 = \mathrm{PatchEmbed}(x) \in \mathbb{R}^{C\times32\times32\times32},
  \label{eq:patchembed}
\end{equation}
\begin{equation}
  z_\ell = \mathrm{SwinStage}_\ell(z_{\ell-1}) \in
  \mathbb{R}^{2^{\ell}C \times d_\ell \times d_\ell \times d_\ell},
  \label{eq:swinstage}
\end{equation}
where $d_\ell = 32/2^{\ell}$ for $\ell=1,\dots,4$, and each
$\mathrm{SwinStage}_\ell$ applies $N_\ell$ transformer blocks
of (shifted-)window multi-head self-attention,
\begin{equation}
  \mathrm{Attn}(Q,K,V) = \mathrm{softmax}\!\left(\frac{QK^\top}{\sqrt{d}} + B\right)V,
  \label{eq:windowattn}
\end{equation}
with relative position bias $B$, a residual MLP sublayer, and a
terminal $2\times$ patch-merging downsample ($z_0,\dots,z_4$ are the
``hierarchical sequence of 5 feature maps'' of
Section~\ref{sec:voxeltower}). The deepest stage
$z_4\in\mathbb{R}^{16C\times2\times2\times2}$ (384 channels at
$2^3$ spatial resolution for $C=24$) is global-average-pooled over its
spatial axes and linearly projected to the shared embedding space:
\begin{equation}
  v = W_V\,\mathrm{GAP}(z_4) + b_V \in \mathbb{R}^{512}, \qquad
  \hat v = v / \lVert v \rVert_2.
  \label{eq:voxelembed}
\end{equation}
All Voxel Encoder parameters $\theta_V$ (patch embedding, all four
stages, and the projection head) are fully fine-tuned
(Section~\ref{sec:voxeltower}).

\paragraph{Finding Encoder $f_T$.}
ClinicalBERT (12 transformer layers, hidden size 768) encodes the
tokenized finding text,
\begin{equation}
  h = \mathrm{ClinicalBERT}_{\theta_T^{(0)}+\Delta\theta_T}(w_{1:L}) \in \mathbb{R}^{L\times768},
  \label{eq:clinicalbert}
\end{equation}
where $\theta_T^{(0)}$ denotes the frozen pretrained weights and
$\Delta\theta_T$ the LoRA adapters. Each query/key/value projection
$W_0\in\mathbb{R}^{d\times k}$ of every self-attention layer is
LoRA-reparameterized,
\begin{equation}
  W_0\,x \;\longrightarrow\; W_0\,x + \frac{\alpha}{r}\,B A\,x,
  \label{eq:lora}
\end{equation}
where $A\in\mathbb{R}^{r\times k}$ and $B\in\mathbb{R}^{d\times r}$ (rank
$r=16$, scale $\alpha=32$), with only
$\Delta\theta_T=\{A_\ell,B_\ell\}_{\ell=1}^{12}$ trainable,
versus 3.68M trainable parameters for the fully fine-tuned Voxel
Encoder (Section~\ref{sec:voxeltower}). The \texttt{[CLS]} hidden state
$h_{\mathrm{CLS}}=h_1$ is linearly projected and normalized:
\begin{equation}
  u = W_T\,h_{\mathrm{CLS}} + b_T \in \mathbb{R}^{512}, \qquad
  \hat u = u / \lVert u \rVert_2.
  \label{eq:textembed}
\end{equation}

\paragraph{Joint similarity and training objective.}
For a batch $\{(x_i,t_i)\}_{i=1}^N$ with embeddings
$\hat v_i = f_V(x_i;\theta_V)$ (Eq.~\ref{eq:voxelembed}) and
$\hat u_i = f_T(t_i;\theta_T^{(0)},\Delta\theta_T)$
(Eq.~\ref{eq:textembed}), a learnable temperature
$\tau=\exp(-s)$ ($s=$~\texttt{logit\_scale}, initialized to
$\log(1/0.07)$) scales the pairwise similarity matrix
\begin{equation}
  S_{ij} = \frac{\hat v_i^\top \hat u_j}{\tau}, \qquad i,j=1,\dots,N,
  \label{eq:simmatrix}
\end{equation}
trained under the symmetric InfoNCE loss
\begin{equation}
  \mathcal{L} = -\frac{1}{2N}\sum_{i=1}^{N}
  \left[ \log\frac{e^{S_{ii}}}{\sum_{j=1}^{N} e^{S_{ij}}}
       + \log\frac{e^{S_{ii}}}{\sum_{j=1}^{N} e^{S_{ji}}} \right],
  \label{eq:infonce}
\end{equation}
with $\theta_V$, $\Delta\theta_T$, and $s$ jointly updated by AdamW
under a cosine learning-rate schedule (Table~\ref{tab:hparams}).

\paragraph{Output: downstream decision rules.}
At inference, the trained embeddings $(\hat v, \hat u)$ realize this
MVP's four downstream outputs as closed-form functions of cosine
similarity (full operational definitions in
Section~\ref{sec:evaluation}): \emph{zero-shot structure
classification} assigns
$\hat y(x) = \arg\max_{c \in \mathcal{C}} \hat v(x)^\top \hat u(\mathrm{prompt}_c)$
over candidate-structure text prompts $\mathcal{C}$;
\emph{bidirectional retrieval} ranks the opposite modality's corpus by
$\hat v_i^\top \hat u_j$ and reports Recall@$\{1,5,10\}$;
\emph{voxel anomaly scoring} (Eq.~\ref{eq:voxelanomaly}) contrasts a
query volume's cosine similarity to normal- vs.\ abnormal-prompt
embeddings; and \emph{diagnosis-reason retrieval} returns the top-$k$
training-corpus finding texts ranked by $\hat v(x)^\top \hat u(t)$.
Thus the full pipeline maps a raw $64^3$ dual-channel voxel (and, at
training time, its paired finding text) to a single 512-d embedding via
Eqs.~\ref{eq:patchembed}--\ref{eq:textembed}, trains that mapping via
Eq.~\ref{eq:infonce}, and every reported downstream metric is a
deterministic function of cosine similarities between such embeddings.

\subsection{Dataset Construction}
\label{sec:dataset}

\subsubsection{Structure Selection}

We query the Allen Institute's \texttt{aics/hipsc\_single\_cell\_image\_
dataset} quilt3 package (\texttt{s3://allencell}) for four structures
spanning distinct organelle/structural categories: \texttt{LMNB1} (nuclear
envelope, lamin B1), \texttt{TOMM20} (mitochondria), \texttt{TUBA1B}
(microtubules), and \texttt{TJP1} (tight junctions, ZO-1). The originally
planned candidate structures included ``TOM20'' (corrected here
to the real HGNC symbol \texttt{TOMM20}) and a plasma-membrane line
``CAAX''; we verified empirically against the dataset's own
\texttt{metadata.csv} that no \texttt{CAAX} structure is present in this
particular quilt3 package, and substituted \texttt{TJP1} as the closest
available membrane-associated analog (Section~\ref{sec:lessons}).

\subsubsection{Per-Cell Volume Extraction}

Each sampled cell's pre-segmented, pre-cropped single-cell volume
(\texttt{crop\_raw}, an OME-TIFF) is fetched via quilt3 and read with
\texttt{aicsimageio}. We empirically determined (Section~\ref{sec:lessons})
that \texttt{crop\_raw} always uses a fixed 3-channel order -- channel 0:
DNA dye, channel 1: cell-membrane dye, channel 2: the structure protein
tagged in that specific cell line -- regardless of cell line, and use
channels 0 and 2 (discarding the membrane channel for this version). Each
channel is percentile-clipped (0.5--99.5\%), min-max normalized to
$[0,1]$, resized via \texttt{scipy.ndimage.zoom} to a fixed
64$\times$64$\times$64 voxel shape (satisfying \texttt{SwinUNETR}'s
patch-size$^5$=32 divisibility constraint), and cached as a
float16 \texttt{.npy} array to avoid repeated S3 fetches and TIFF decoding
during training.

\subsubsection{Mitotic Oversampling and Proxy Labels}

Each cell carries a \texttt{cell\_stage} annotation: \texttt{M0}
(interphase) or one of five combined mitotic sub-stage buckets
(\texttt{M1M2}, \texttt{M3}, \texttt{M4M5}, \texttt{M6M7\_single},
\texttt{M6M7\_complete}). Because mitosis is naturally rare (we measured
2--13\% of cells per structure, depending on structure), we deliberately
oversample mitotic cells up to 20\% of each structure's 750-cell quota
(topped up with interphase cells for the remainder), so that the
anomaly-scoring ROC-AUC validation (Section~\ref{sec:scoring}) has enough
positive (abnormal-like) samples. \texttt{cell\_stage}\,$=$\,\texttt{M0}
is treated as normal-like and all other values as abnormal-like
(\texttt{cell\_anomaly\_proxy\_labels\_voxel.yaml}), the same ``real but
proxy'' labeling convention as v0.1.0/v0.2.0's weak-label files: a formal,
instrument-derived ground truth for cell division state, but not a
literal analog of pavement-material damage.

\subsubsection{Template Finding-Text Generation}

No free-text pathology report accompanies this dataset, so each cell's
finding text is rendered from a small ensemble of templates keyed by its
interphase/mitotic bucket (e.g.\ \emph{``a 3D confocal fluorescence volume
of an interphase human iPS cell showing an intact nuclear envelope (lamin
B1)''} vs.\ \emph{``a 3D microscopy volume showing mitochondria (TOM20)
undergoing mitotic reorganization (stage M4M5)''}), the same
template-ensemble convention as v0.1.0/v0.2.0's caption generation,
deliberately simpler than an LLM-expanded alternative considered and
rejected for this version (Section~\ref{sec:limitations}).

\subsubsection{Splits and Progressive Scales}

We hold out a fixed 125-cell/structure test split (500 cells total) and
pool the remaining 625 cells/structure (2{,}500 total) as the progressive
training corpus. Unlike v0.1.0/v0.2.0's four-stage 1k/2k/3k/4k protocol,
we use three stages -- 0.8k/1.6k/2.5k -- stratified by
\texttt{structure\_name}, reflecting the substantially higher per-sample
compute cost of 3D volumes relative to 2D images.

\subsection{Contrastive Domain Adaptation}
\label{sec:contrastive}

Table~\ref{tab:hparams} summarizes training hyperparameters, confirmed via
a GPU-memory smoke test (Section~\ref{sec:smoketest}) before any data
download or real training began.

\begin{table}[h]
\centering
\caption{VoxelSwinCLAIP training hyperparameters.}
\label{tab:hparams}
\footnotesize
\setlength{\tabcolsep}{4pt}
\begin{tabular}{ll}
\toprule
\textbf{Parameter} & \textbf{Value} \\
\midrule
Voxel shape / channels & 64$^3$ / [DNA, structure] \\
3D Swin feature size & 24 (divisible by 12) \\
Voxel Encoder adaptation & full fine-tune \\
Text LoRA $r$ / $\alpha$ & 16 / 32 \\
Shared embedding dim & 512 \\
Batch size & 12 \\
Precision & bf16 autocast \\
Gradient checkpointing & enabled \\
Learning rate & $1\times10^{-4}$ \\
Epochs & 15 \\
Optimizer & AdamW, cosine schedule \\
\midrule
GPU & RTX 4060 Ti 16GB \\
CUDA & 12.4 \\
PyTorch / MONAI & 2.6.0+cu124 / 1.5.2 \\
\bottomrule
\end{tabular}
\end{table}

\subsubsection{GPU Feasibility Smoke Test}
\label{sec:smoketest}

Because this is the first 3D-voxel architecture in this MVP progression,
we validated the 16GB VRAM budget empirically \emph{before} any data
download, via a dummy forward/backward step at several batch sizes:
3.26\,GB (batch\,4), 8.68\,GB (batch\,12), and 11.39\,GB (batch\,16) peak
CUDA memory. We selected batch\,size\,12 (8.68\,GB, $\approx$54\% of
budget) for a comfortable safety margin across many training epochs while
maximizing in-batch negatives for InfoNCE.

\subsection{Evaluation Protocol}
\label{sec:evaluation}

We evaluate four downstream capabilities on the held-out 500-cell
(125/structure) test split, comparing a freshly initialized
\texttt{VoxelSwinCLAIP} (random 3D Swin + pretrained-but-not-contrastively-
trained ClinicalBERT, \texttt{base}) against each progressive-scale
fine-tuned checkpoint:

\textbf{(1) Zero-shot structure classification}: prompt template ``a 3D
confocal fluorescence volume showing \{c\}'', 4-way top-1 accuracy
(chance = 25\%). As in v0.2.0, no external literature zero-shot baseline
exists for this architecture/domain, so \texttt{base} accuracy is a
self-baseline reference only.

\textbf{(2) Bidirectional retrieval}: voxel$\to$text and text$\to$voxel
Recall@\{1,5,10\}.

\textbf{(3) Voxel anomaly scoring}: analogous to v0.2.0's compositional
cell-anomaly score, with ``normal'' (interphase) and
``abnormal'' (mitotic) text prototypes built per structure from
\texttt{scoring\_rules\_voxel.yaml}, validated by ROC-AUC against
\texttt{cell\_anomaly\_proxy\_labels\_voxel.yaml}'s
\texttt{cell\_stage}-based grouping.
\begin{equation}
  \text{voxel\_anomaly}(x) = \cos(x, p_{\mathrm{norm}}) - \cos(x, p_{\mathrm{abn}}),
  \label{eq:voxelanomaly}
\end{equation}
\label{sec:scoring}

\textbf{(4) Diagnosis-reason retrieval} (new in v0.3.0): for each
abnormal-like (mitotic) test volume, we retrieve the nearest-neighbor
finding text from the pooled train+val corpus by voxel-to-text cosine
similarity, and report hit-rate@\{1,5\}: whether any of the top-$k$
retrieved findings shares both the query's \texttt{structure\_name} and
its interphase-vs-mitotic state. This directly operationalizes this MVP
series' originally envisioned ``abnormality diagnosis reason search''
output within a retrieval-only scope (free-text generation is deferred to v0.4.0,
Section~\ref{sec:limitations}).

\section{Implementation Lessons: Metadata, Architecture, and Adaptation Strategy}
\label{sec:lessons}

Reusing a real biomedical imaging dataset's metadata at face value
surfaced two problems that a careful implementation-time verification
step caught before they could silently corrupt the training corpus.

\subsection{Metadata Column Scope Mismatch: FOV-Level vs.\ Crop-Level Channels}

Our initial design assumed that \texttt{metadata.csv}'s per-cell
\texttt{ChannelNumberStruct}/\texttt{ChannelNumber405} columns directly
index the channel axis of the per-cell \texttt{crop\_raw} volume. A first
real download run failed immediately with
\texttt{IndexError: index 5 is out of bounds for axis 0 with size 3}: we
confirmed, by downloading one cell and inspecting its OME metadata
directly, that \texttt{crop\_raw} always has exactly 3 channels in a
\emph{fixed} order (DNA, membrane, structure) for every cell line, and
that the \texttt{ChannelNumber*} columns instead index the much larger
per-field-of-view (FOV) multi-channel image, not the already
channel-reduced single-cell crop. We corrected this to hardcoded indices
(0 = DNA, 2 = structure) rather than per-row metadata lookup, and verified
across multiple structures that this eliminated all channel-related
failures.

\subsection{Verifying Candidate Structures Against the Real Dataset}

The originally planned candidate structure list (nucleus, mitochondria,
microtubule, plus a plasma-membrane line named ``CAAX'') was drawn from
general knowledge of the Allen Cell Collection rather than this specific
quilt3 package's actual contents. Querying the real
\texttt{structure\_name} column revealed 25 available structures, none
named \texttt{CAAX}; the closest available membrane-associated analog,
\texttt{TJP1} (tight junctions), was substituted instead
(Section~\ref{sec:dataset}). We also corrected ``TOM20'' to its real HGNC
gene symbol, \texttt{TOMM20}. Both corrections are intentionally
documented here rather than silently fixed, to illustrate that even a
reasonable, literature-informed data-source candidate list should be
verified against the actual target package before implementation,
particularly for datasets whose metadata.csv schema may itself differ
across package versions (we also found the \texttt{cell\_stage} column
uses combined mitotic-substage buckets, e.g.\ \texttt{M6M7\_single}, rather
than the individual \texttt{M1}--\texttt{M7} labels shown in the
dataset's own older tutorial notebooks).

\subsection{LoRA-Only Adaptation Collapses on a Randomly Initialized Backbone}
\label{sec:lessons2}

A first complete 3-scale training run (15 epochs each, LoRA rank $r=8$ on
only the window-attention \texttt{qkv}/\texttt{proj} projections, all
other Voxel Encoder parameters frozen at random initialization) completed
without error and showed a healthy, monotonically decreasing InfoNCE loss
(2.5k scale: 1.83$\to$1.16), yet every downstream evaluation -- zero-shot
accuracy, bidirectional retrieval, and anomaly scoring -- remained exactly
at chance for every scale including the best-trained (2.5k) checkpoint.
Inspecting voxel embeddings directly revealed the cause: pairwise cosine
similarity across 8 distinct-structure test volumes ranged 0.93--0.99 for
every pair, base model and all three trained scales alike -- the Voxel
Encoder was producing near-input-invariant embeddings regardless of the
actual cell structure. This is a direct consequence of applying LoRA
(designed to \emph{adapt} an already-informative pretrained prior with a
small low-rank delta) to a backbone with \emph{no} pretrained prior at
all: with \texttt{patch\_embed}'s convolution and every
\texttt{PatchMerging} downsampling layer frozen at their random
initialization, a rank-8 delta on only the attention projections cannot
compensate for an otherwise-uninformative random computation graph. We
switched the Voxel Encoder to full fine-tuning (all parameters trainable,
Section~\ref{sec:voxeltower}) and retrained all three scales; pairwise
voxel-embedding cosine similarity on the same 8-volume probe now shows a
clear separation (same-structure pairs 0.98--0.997, cross-structure pairs
0.79--0.93), and all four downstream evaluations move measurably away
from chance (Section~\ref{sec:results}). We record this as a general
lesson for any future contrastive vision-language work that pairs a
\emph{non-pretrained} modality tower with LoRA: LoRA's parameter
efficiency assumption only holds when the frozen backbone already encodes
useful structure.

\section{Experimental Setup}
\label{sec:setup}

All experiments run on a single RTX 4060 Ti 16GB GPU (CUDA 12.4,
PyTorch 2.6.0, MONAI 1.5.2, peft 0.21.2, transformers 5.22.1). Data
preparation, progressive splitting, training (3 scales), and all four
evaluation protocols are implemented as standalone scripts mirroring
v0.1.0/v0.2.0's pipeline structure, each responsible for one stage of the
pipeline:
\begin{itemize}[leftmargin=*,itemsep=1pt]
  \item \textbf{Dataset preparation}: fetches and preprocesses per-cell
        volumes from the Allen Institute source (channel selection,
        normalization, resizing; Section~\ref{sec:dataset}).
  \item \textbf{Progressive split generation}: builds the fixed test
        holdout and the stratified 0.8k/1.6k/2.5k progressive train+val
        pools.
  \item \textbf{Contrastive training}: trains \texttt{VoxelSwinCLAIP}
        under the symmetric InfoNCE objective at each progressive scale.
  \item \textbf{Zero-shot evaluation}: computes 4-way zero-shot
        structure classification accuracy.
  \item \textbf{Retrieval evaluation}: computes bidirectional
        voxel$\leftrightarrow$text Recall@\{1,5,10\}.
  \item \textbf{Anomaly scoring}: computes the compositional-prompt
        voxel anomaly score and its ROC-AUC.
  \item \textbf{Diagnosis-reason retrieval}: performs nearest-neighbor
        finding-text retrieval and reports hit-rate@\{1,5\}.
  \item \textbf{Figure generation}: renders all result figures and
        tables from the logged evaluation outputs.
\end{itemize}

Dataset acquisition (quilt3, 12 concurrent download workers) processed all
4 structures with \emph{zero} unprocessable/corrupted cells after the
channel-indexing fix (Section~\ref{sec:lessons}), yielding the full target
corpus: 3{,}000 cells (750/structure: 150 mitotic + 600 interphase each,
20\% oversampling as configured), split into 500 test (125/structure) and
2{,}500 train+val (625/structure, further split into 1{,}250/1{,}250 for
the fixed train/val pools used by progressive split generation). The
test split's realized abnormal-like (mitotic) count is
114/500 (22.8\%), close to but not identical to the 25\% targeted overall
by the 20\%-of-750 oversampling rule, since the fixed 125/structure test
holdout is drawn independently of the mitotic/interphase split ratio.

\section{Results}
\label{sec:results}

\subsection{Zero-Shot Structure Classification}

\begin{figure}[h]
\centering
\includegraphics[width=0.47\textwidth]{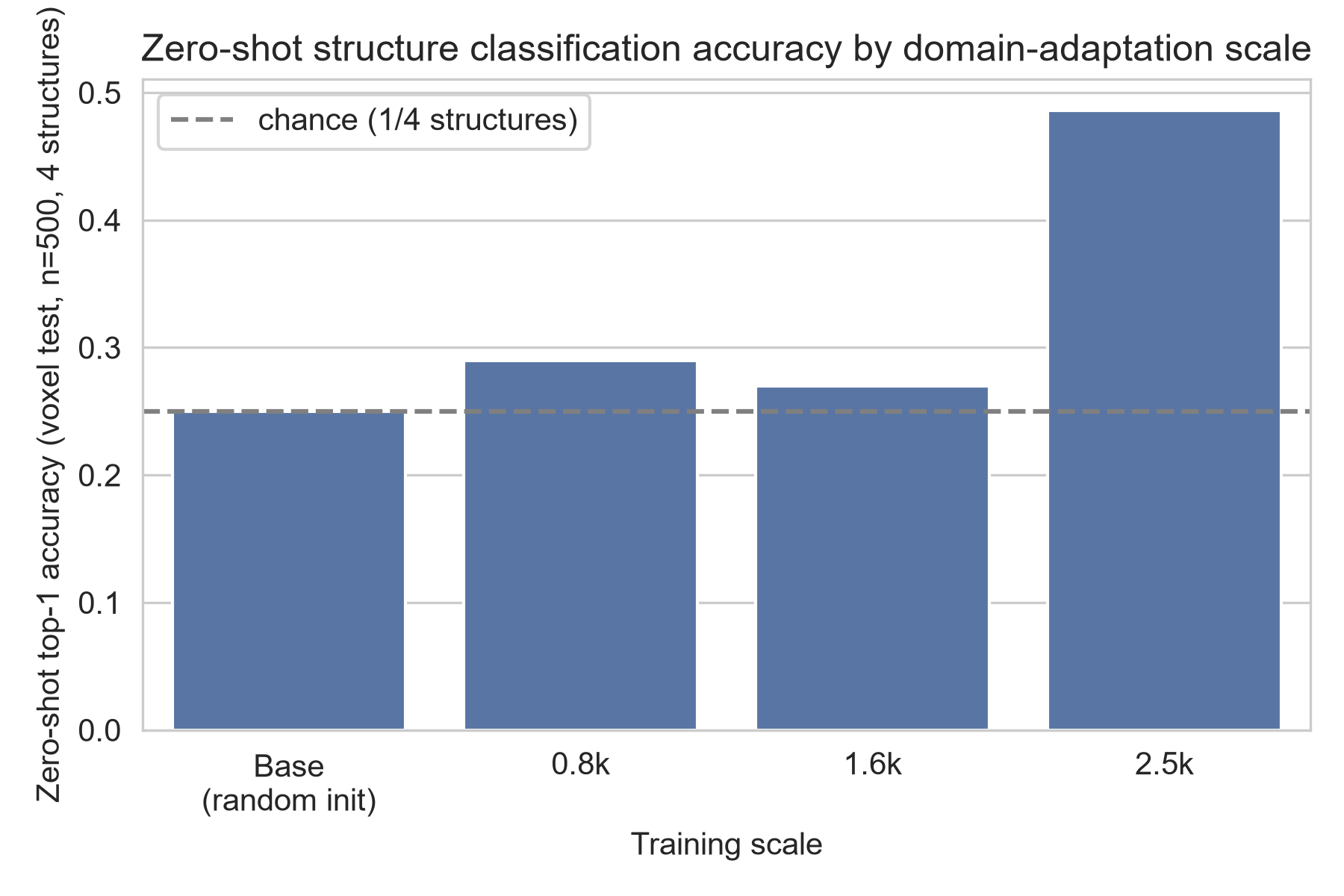}
\caption{Zero-shot 4-way structure classification accuracy by
domain-adaptation scale, against the 25\% chance line.}
\label{fig:zeroshot}
\end{figure}

Table~\ref{tab:zeroshot} and Figure~\ref{fig:zeroshot} report 4-way
zero-shot structure classification accuracy. Accuracy rises
non-monotonically with scale -- 25.0\% (base, exactly chance) to 29.0\%
(0.8k), a dip to 27.0\% (1.6k), then a clear jump to 48.6\% (2.5k, nearly
double chance) -- indicating that useful structure-discriminative
representation emerges only with the full 2{,}500-cell training pool
under this fully-fine-tuned (not LoRA-only) Voxel Encoder.

\begin{table}[h]
\centering
\caption{Zero-shot 4-way structure classification accuracy (voxel test,
n=500; chance=25\%).}
\label{tab:zeroshot}
\begin{tabular}{lr}
\toprule
\textbf{Scale} & \textbf{Accuracy} \\
\midrule
Base (random init)       & 25.0\% \\
0.8k                     & 29.0\% \\
1.6k                     & 27.0\% \\
2.5k                     & 48.6\% \\
\bottomrule
\end{tabular}
\end{table}

\subsection{Progressive Training and Validation Loss}

\begin{figure}[h]
\centering
\includegraphics[width=0.47\textwidth]{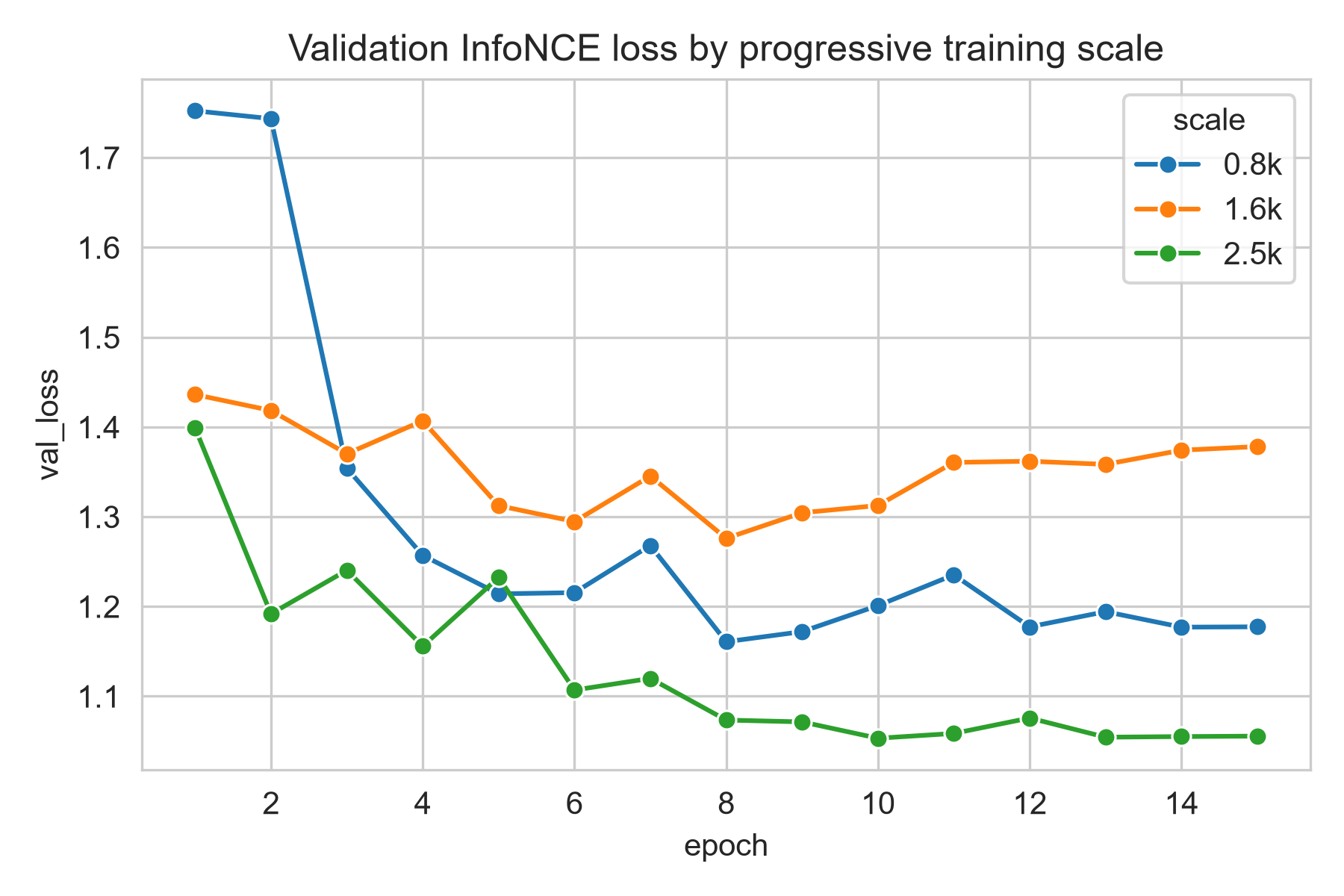}
\caption{Validation InfoNCE loss across training epochs, by progressive
scale.}
\label{fig:valloss}
\end{figure}

Table~\ref{tab:valloss} reports final-epoch (epoch 15) validation InfoNCE
loss, training-pool size, and wall-clock duration for each scale
(Figure~\ref{fig:valloss}). Loss decreases from 0.8k to 2.5k overall
(1.177$\to$1.056) but is non-monotonic at 1.6k (1.378, higher than both
neighbors) -- consistent with the zero-shot accuracy dip at the same
scale (Table~\ref{tab:zeroshot}) -- suggesting the 1.6k stratified subset
may have been a comparatively harder draw for this run's fixed random
seed. Peak training memory is stable at 9.40--9.93\,GB across all three
scales, comfortably within the 16GB budget and close to the Phase 0 smoke
test's 8.68\,GB estimate (the modest increase reflects the switch
to full voxel-backbone fine-tuning, Section~\ref{sec:lessons2}).

\begin{table}[h]
\centering
\caption{Progressive contrastive training results (full voxel-backbone
fine-tuning + ClinicalBERT LoRA).}
\label{tab:valloss}
\footnotesize
\setlength{\tabcolsep}{3pt}
\begin{tabular}{lrrrr}
\toprule
\textbf{Scale} & \textbf{$n_{\text{train}}$} & \textbf{Val.\ loss} & \textbf{Duration} & \textbf{Peak mem} \\
\midrule
0.8k & 680   & 1.177 & 14.6\,min & 9.40\,GB \\
1.6k & 1{,}360 & 1.378 & 29.0\,min & 9.93\,GB \\
2.5k & 2{,}125 & 1.056 & 45.4\,min & 9.93\,GB \\
\bottomrule
\end{tabular}
\end{table}

\subsection{Bidirectional Retrieval}

\begin{figure}[h]
\centering
\includegraphics[width=0.47\textwidth]{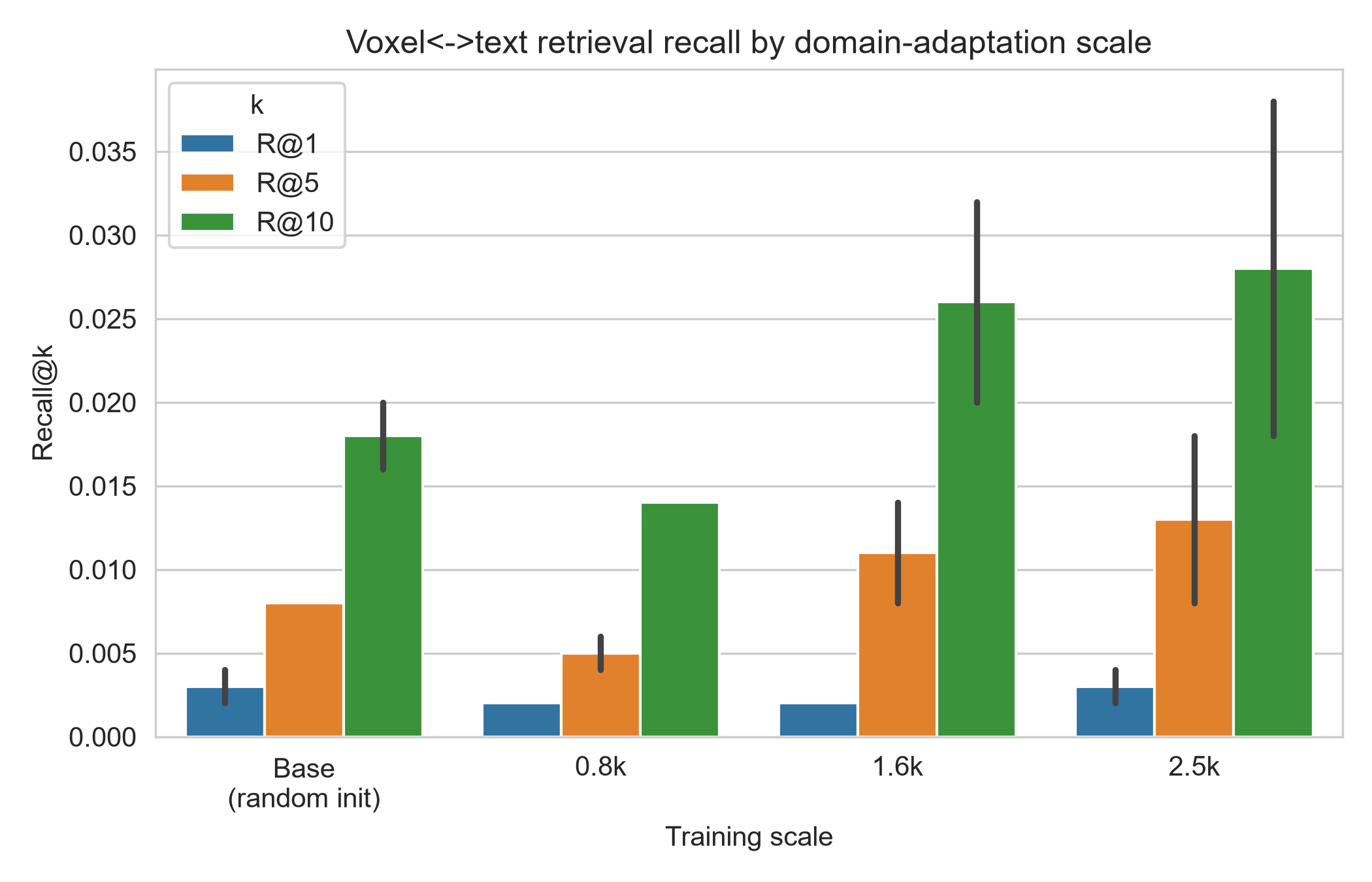}
\caption{Bidirectional voxel$\leftrightarrow$text retrieval recall by
domain-adaptation scale.}
\label{fig:retrieval}
\end{figure}

Figure~\ref{fig:retrieval} reports voxel$\to$text and text$\to$voxel
Recall@\{1,5,10\} on the test split. Recall remains low and close to base
across all scales: voxel$\to$text R@10 ranges 1.4--2.0\% (base 2.0\%) and
text$\to$voxel R@10 ranges 1.4--3.8\% (base 1.6\%), with only a mild
upward trend in text$\to$voxel R@10 from base to 2.5k (1.6\%$\to$3.8\%).
This mirrors v0.2.0's finding that retrieval recall is a weak signal when
only a handful of classes and caption templates are available: with only
4 structures $\times$ 2 cell-stage buckets $\times$ a small template
ensemble, very many test cells share a near-identical or literally
identical finding text, making the single-correct-match retrieval task
severely ill-posed regardless of embedding quality
(Section~\ref{sec:limitations}).

\subsection{Voxel Anomaly Scoring}
\label{sec:results_scoring}

\begin{figure}[h]
\centering
\includegraphics[width=0.47\textwidth]{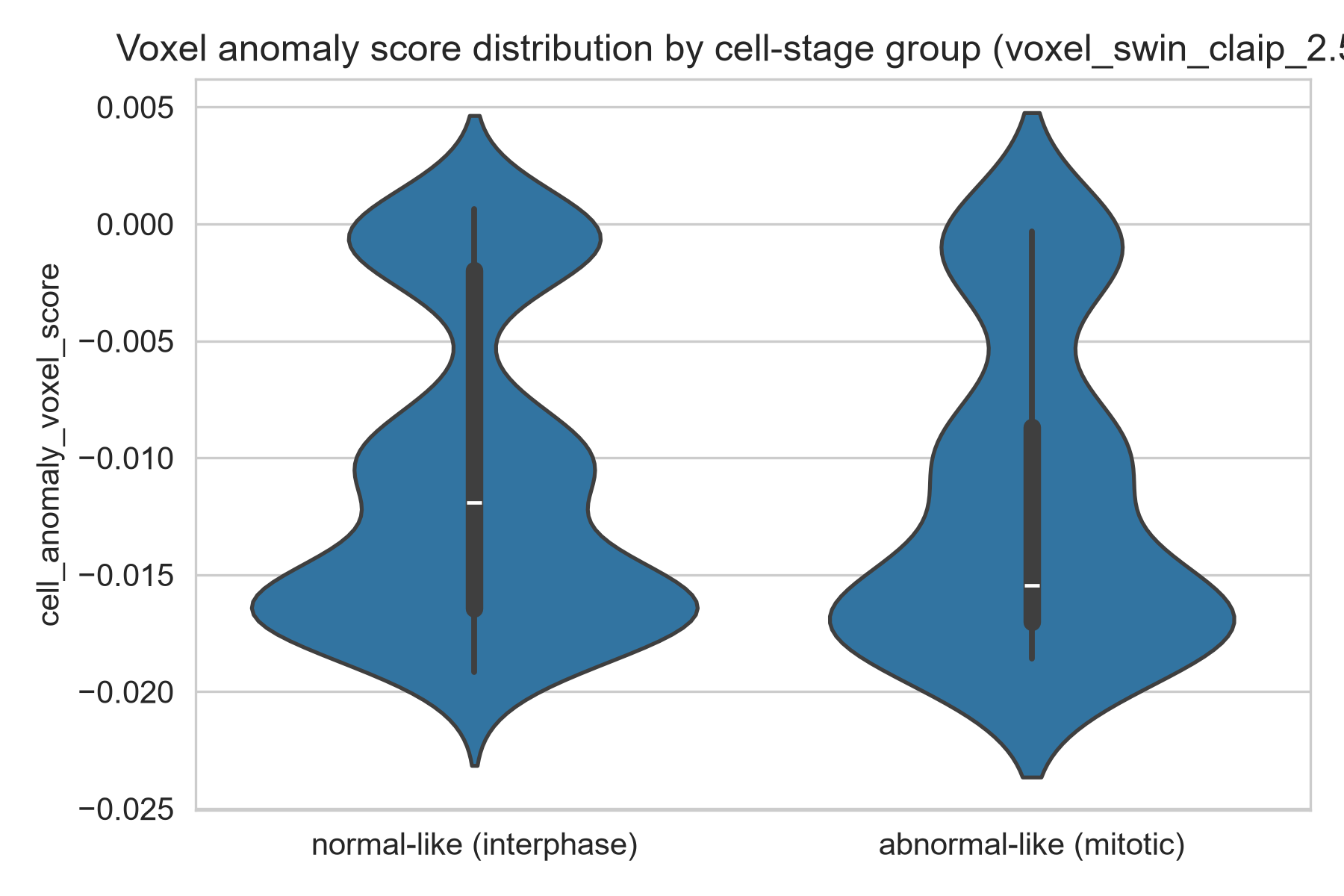}
\caption{Voxel anomaly score distribution by cell-stage group (2.5k
model): abnormal-like (mitotic) vs.\ normal-like (interphase).}
\label{fig:violin}
\end{figure}

Table~\ref{tab:anomaly} reports ROC-AUC between the compositional voxel-
anomaly score and the \texttt{cell\_stage}-based abnormal-like/normal-like
grouping (114/500 abnormal-like test cells throughout, reported
transparently for completeness). AUC rises from 0.486 (base, at chance)
to 0.628 (1.6k) before falling back to 0.567 (2.5k) -- none of the
domain-adapted scales clear the provisional 0.70 sanity threshold used in
v0.1.0/v0.2.0, unlike those versions' 2D results. Figure~\ref{fig:violin}
compares the base and 2.5k score distributions by cell-stage group.

\begin{table}[h]
\centering
\caption{Voxel anomaly scoring ROC-AUC against the cell\_stage-based
abnormal-like/normal-like proxy grouping (n=500, 114 abnormal-like;
threshold 0.70).}
\label{tab:anomaly}
\footnotesize
\setlength{\tabcolsep}{4pt}
\begin{tabular}{lrl}
\toprule
\textbf{Model} & \textbf{ROC-AUC} & \textbf{Status} \\
\midrule
Base & 0.486 & BELOW\_THRESHOLD \\
0.8k & 0.488 & BELOW\_THRESHOLD \\
1.6k & 0.628 & BELOW\_THRESHOLD \\
2.5k & 0.567 & BELOW\_THRESHOLD \\
\bottomrule
\end{tabular}
\end{table}

\subsection{Diagnosis-Reason Retrieval (New in v0.3.0)}

Table~\ref{tab:diagretrieval} reports hit-rate@\{1,5\} for the 114
abnormal-like test queries against the 2{,}500-cell train+val corpus.
Unlike the other three metrics, this result is sharply
non-monotonic and binary-looking across scales: base, 0.8k, and 2.5k all
score exactly 0.0 (no query's top-5 nearest-neighbor findings matched
both its structure and mitotic state), while 1.6k reaches 29.8\% at both
$k=1$ and $k=5$ simultaneously -- the equality between hit-rate@1 and
hit-rate@5 indicates that whenever the top-1 retrieval was correct for
this checkpoint, no \emph{additional} correct matches appeared at ranks
2--5 (and vice versa), consistent with a corpus in which same-
structure/same-stage findings cluster tightly in embedding space once a
model finds that cluster at all.

\begin{table}[h]
\centering
\caption{Diagnosis-reason retrieval hit-rate@k (n=114 abnormal-like test
queries, 2,500-cell corpus).}
\label{tab:diagretrieval}
\begin{tabular}{lrr}
\toprule
\textbf{Model} & \textbf{Hit-rate@1} & \textbf{Hit-rate@5} \\
\midrule
Base & 0.0\%  & 0.0\% \\
0.8k & 0.0\%  & 0.0\% \\
1.6k & 29.8\% & 29.8\% \\
2.5k & 0.0\%  & 0.0\% \\
\bottomrule
\end{tabular}
\end{table}

\subsection{Inference Cost on the 16GB eGPU}

Peak training memory ranged 9.40--9.93\,GB across the three scales
(bf16, batch\_size=12, full voxel-backbone fine-tuning + ClinicalBERT
LoRA), comfortably within the 16GB budget (Table~\ref{tab:valloss}),
confirming the Phase 0 smoke test's pre-registered estimate remained
accurate after the architecture correction of Section~\ref{sec:lessons2}.

\section{Discussion}
\label{sec:discussion}

\subsection{The 2D Recipe Does Not Transfer Unchanged -- It Requires an Adaptation-Strategy Correction}

Naively porting v0.1.0/v0.2.0's ``freeze backbone, LoRA the attention
projections'' recipe to a 3D Swin Transformer produced a complete failure
mode invisible from the training loss alone: loss decreased healthily
while every downstream metric stayed at chance, because the frozen-random
Voxel Encoder produced near-input-invariant embeddings
(Section~\ref{sec:lessons2}). This is a qualitatively different failure
from anything observed in v0.1.0/v0.2.0, where both CLIP towers already
carried strong pretrained priors that LoRA could meaningfully steer. Once
corrected to full fine-tuning, the architecture \emph{does} support the
same symmetric-InfoNCE training loop and progressive-scale protocol
without further changes, and zero-shot structure accuracy reaches 48.6\%
at 2.5k (Table~\ref{tab:zeroshot}) -- a real, if partial, transfer of the
methodology. The key lesson is not that the 2D recipe fails to generalize
architecturally, but that its \emph{LoRA-vs-full-fine-tune} choice must be
re-decided per tower based on whether that tower starts from a pretrained
prior, not inherited by convention from the prior MVPs.

\subsection{A Metric Dissociation: Structure Identity Is Learned; Cell-Cycle State Is Not (Yet)}

Zero-shot structure accuracy improves substantially with scale (25.0\%
$\to$ 48.6\%), but voxel anomaly-scoring AUC against the mitotic/
interphase proxy label peaks at a much more modest 0.628 (1.6k) and never
clears the 0.70 sanity threshold used in v0.1.0/v0.2.0
(Table~\ref{tab:anomaly}). This dissociation suggests the InfoNCE
objective, driven primarily by the easier and more class-balanced
structure-identity signal (4 roughly-equal-sized classes vs.\ an 114/500
minority abnormal-like class), learns to separate structures in voxel
embedding space more readily than it learns the subtler mitotic-vs-
interphase morphological difference \emph{within} a structure -- despite
our deliberate 20\% mitotic oversampling (Section~\ref{sec:dataset}).
Diagnosis-reason retrieval's sharp, nearly all-or-nothing pattern across
scales (0.0\% at base/0.8k/2.5k, 29.8\% at 1.6k;
Table~\ref{tab:diagretrieval}) is consistent with this: it requires
\emph{both} signals to align correctly for the same test cell, a strictly
harder joint criterion than either signal alone, which with only 114 test
queries is also highly sensitive to small differences in exactly which
mitotic sub-population each checkpoint happens to represent well.

\subsection{Mitotic Oversampling Alone Did Not Guarantee a Stable Anomaly Signal}

Our 20\% mitotic oversampling target (Section~\ref{sec:dataset}) succeeded
at its narrow goal -- 114/500 (22.8\%) of the test split is abnormal-like,
comfortably enough samples for ROC-AUC to be computable and meaningful
(unlike the ``INSUFFICIENT\_POSITIVE\_SAMPLES'' guard built into
\texttt{scoring\_cell\_anomaly\_voxel.py} for the case where it would not
be) -- but did not by itself guarantee that the \emph{learned}
representation would separate interphase from mitotic cells well; AUC
remained below the 0.70 threshold at every scale. This suggests that, for
mitotic state specifically, the compositional-prompt scoring approach
inherited from v0.1.0/v0.2.0 may need either more mitotic-specific
template diversity, more training epochs targeted at this sub-task, or
(Section~\ref{sec:limitations}) a genuinely differently weighted training
objective that does not let the easier structure-identity signal dominate.

\subsection{Diagnosis-Reason Retrieval as a Minimal, and Currently Fragile, ``Explainability'' Step}

The 1.6k checkpoint's 29.8\% hit-rate@\{1,5\} shows that nearest-neighbor
retrieval over template-generated finding text \emph{can} recover a
jointly-correct (structure + cell-stage) explanation for a meaningful
fraction of abnormal-like test cells, operationalizing this MVP series'
``abnormality diagnosis reason search'' goal in a minimal, retrieval-only
form. However, the 0.0\% result at every other scale -- including the
otherwise-best-performing 2.5k checkpoint on zero-shot accuracy -- shows
this capability is not yet a stable, monotonically-improving function of
training scale, and should be read as an existence proof that the
evaluation protocol works, not as evidence that the current
\texttt{VoxelSwinCLAIP} training recipe reliably produces this capability.

\subsection{Limitations}
\label{sec:limitations}

\textbf{Proxy dataset, not production data.}
As in v0.1.0/v0.2.0, results reported here validate the \emph{methodology}
-- 3D voxel/text dual-encoder contrastive training, evaluation, and
scoring -- not material-durability prediction accuracy on real
road-pavement 3D microscope data, which remains to be established once
production imagery and inspection labels are collected.

\textbf{Template, not real, finding text.}
Unlike a genuine Medical CLAIP deployment, our ``finding text'' is
template-generated from structured metadata (\texttt{structure\_name},
\texttt{cell\_stage}), not a free-text pathology report, even though the
text encoder (ClinicalBERT) is pretrained on real clinical notes. This was
a deliberate MVP-scope decision (Section~\ref{sec:dataset}) to avoid the
added cost and quality-verification burden of LLM-expanded captions;
future versions should revisit this once real pavement-inspection reports
are available.

\textbf{Two-tower, not three-tower, contrastive learning.}
The full, long-term ``Medical CLAIP'' design (Figure~\ref{fig:architecture})
adds a third Gene/
Structure-Identity Encoder trained jointly with the Voxel and Finding
Encoders. This version implements only the Voxel$\leftrightarrow$Finding
two-tower contrastive objective; the third tower is deferred to v0.4.0 so
that the new 3D architecture's training/evaluation pipeline could first be
validated in isolation.

\textbf{Mitotic cell rarity.}
Despite deliberate 20\% oversampling, mitosis remains a naturally
short-lived cell state (2--13\% of cells per structure before
oversampling); the resulting abnormal-like sample counts, while enriched,
are still smaller than the normal-like class, and any ROC-AUC reported in
Section~\ref{sec:results} should be read alongside the actual $n$ reported
in \texttt{scoring\_cell\_anomaly\_voxel.py}'s output JSON.

\textbf{Anomaly scoring did not clear the provisional threshold.}
Unlike v0.1.0/v0.2.0, where domain adaptation pushed proxy-label ROC-AUC
well past the 0.70 sanity bar, every v0.3.0 scale remained below it
(peak 0.628 at 1.6k; Table~\ref{tab:anomaly}). We report this as a
genuine negative result for the mitotic/interphase anomaly-scoring task
specifically, not as a failure of the overall methodology (zero-shot
structure accuracy, a different downstream task on the same embeddings,
does show a clear positive scale trend).

\textbf{Full fine-tuning sacrifices parameter efficiency.}
Section~\ref{sec:lessons2}'s correction trades away LoRA's parameter-
efficiency benefit for the Voxel Encoder (3.68M vs.\ $\sim$98K trainable
parameters); this was a necessary and deliberate choice given the
backbone's random initialization, but means this version's Voxel Encoder
adapters are not as lightweight or composable as v0.1.0/v0.2.0's LoRA
adapters, or as this version's own ClinicalBERT LoRA adapter.

\textbf{No externally pretrained 3D Swin weights.}
The Voxel Encoder is randomly initialized rather than loaded from a
CT-pretrained MONAI checkpoint, since the domain gap between CT and
fluorescence confocal microscopy is large and the transfer benefit
unverified; this is a candidate ablation for future work.

\textbf{Dataset licensing.}
The Allen Institute hiPSC dataset is distributed under a non-commercial
research/education license requiring attribution; any production or
commercial reuse must be independently re-verified against the Allen
Institute's current Terms of Use.

\textbf{Hardware.}
All experiments run on a single RTX 4060 Ti 16GB GPU, identical to
v0.1.0/v0.2.0.

\section{Conclusion}
\label{sec:conclusion}

We extended our v0.1.0/v0.2.0 2D contrastive-learning domain-adaptation
methodology to a 3D-voxel, clinical-text architecture
(\texttt{VoxelSwinCLAIP}: a fully fine-tuned 3D Swin Transformer Voxel
Encoder paired with a LoRA-adapted ClinicalBERT Finding Encoder, trained
with the same symmetric InfoNCE objective), trained on a newly assembled
3D biological proxy corpus from the Allen Institute's public hiPSC
single-cell image dataset, and introduced a new diagnosis-reason
retrieval evaluation operationalizing this MVP series' originally
envisioned explainable-diagnosis goal within a retrieval-only MVP scope. Key
findings are as follows.
\begin{enumerate}[leftmargin=*,itemsep=1pt]
  \item \textbf{LoRA's parameter-efficiency assumption does not hold for a
        non-pretrained backbone}: an initial LoRA-only Voxel Encoder
        (attention projections only, backbone otherwise frozen at random
        initialization) trained to a healthy-looking decreasing InfoNCE
        loss while every downstream metric stayed at exact chance, because
        voxel embeddings collapsed to near-input-invariance (pairwise
        cosine similarity 0.93--0.99). Switching to full fine-tuning of
        the Voxel Encoder (while keeping LoRA for the genuinely pretrained
        ClinicalBERT Finding Encoder) resolved this and is the single most
        consequential architectural finding of this MVP
        (Section~\ref{sec:lessons2}).
  \item \textbf{A GPU feasibility smoke test de-risked hyperparameter
        selection before any data download}: empirically measured peak
        VRAM (3.26/8.68/11.39\,GB at batch 4/12/16 for the original
        LoRA-only design) let us confirm batch\,size\,12 as a safe default
        before committing to hours of real S3 downloads and training; the
        corrected full-fine-tune architecture's peak memory
        (9.40--9.93\,GB) remained well within the same budget
        (Section~\ref{sec:smoketest}).
  \item \textbf{Zero-shot structure classification improves substantially
        with scale, but anomaly scoring and retrieval do not (yet)}:
        4-way zero-shot structure accuracy rises from 25.0\% (base,
        chance) to 48.6\% (2.5k), while voxel anomaly-scoring ROC-AUC
        against the mitotic/interphase proxy label peaks at a more modest
        0.628 (1.6k, below the 0.70 sanity threshold) and
        bidirectional retrieval recall stays low across all scales
        (R@10 $\le$3.8\%) -- a metric dissociation suggesting the model
        learns structure identity more readily than the subtler
        cell-cycle-state signal (Section~\ref{sec:discussion}).
  \item \textbf{Diagnosis-reason retrieval is a working but fragile
        capability}: the 1.6k checkpoint reaches 29.8\% hit-rate@\{1,5\}
        for retrieving a jointly-correct (structure + mitotic-state)
        explanation for abnormal-like test cells, while base/0.8k/2.5k
        all score exactly 0.0\%, demonstrating the evaluation protocol
        works but is not yet a stable function of training scale under
        the current recipe (Section~\ref{sec:results}).
  \item \textbf{Real biomedical metadata requires implementation-time
        verification, not just literature-informed design}: two
        independent corrections -- a channel-index metadata scope mismatch
        and a candidate-structure-name mismatch (``CAAX''/``TOM20'') --
        were both caught only by querying the actual dataset at
        implementation time, not by reading documentation alone
        (Section~\ref{sec:lessons}).
  \item \textbf{The full-fine-tuning fix does not generalize unconditionally
        across diversity/scale} (v0.3.1): extending from 4 to 8 structures
        and 3,000 to 8,000 cells, under the identical architecture and
        training recipe that resolved v0.3.0's collapse, produced the
        \emph{same} collapse signature (zero-shot accuracy pinned exactly
        at chance, 12.5\%, at every trained scale up to 4.8k; near-chance
        retrieval; anomaly AUC still below 0.70) despite InfoNCE loss
        falling even lower than in v0.3.0. Critically, this extension
        increased structural diversity and dataset scale
        \emph{simultaneously, in a single step}, confounding which factor
        is responsible; validation loss nonetheless stayed smoothly
        scale-ordered (Figure~\ref{fig:v031valloss}) and diagnosis-reason
        hit-rate stayed above the random-init base at every trained scale
        (Figure~\ref{fig:v031diaghit}), so not every signal collapsed.
        This is the central diversity/generalization finding of this MVP:
        training-loss health is not sufficient evidence that a
        contrastive methodology's \emph{fix} generalizes to a new
        diversity/scale regime, and downstream-metric verification is
        required at each such step (Section~\ref{sec:v031}).
\end{enumerate}

Future work will: (i) diagnose the v0.3.1 collapse directly (per-epoch
embedding-diversity probes, learning-rate/batch-size sweeps, or an
auxiliary structure-classification loss) rather than only detecting it
post hoc; (ii) re-run the diversity/scale extension as a \emph{staged}
curriculum (e.g., 4$\to$6$\to$8 structures, each first validated at the
original 3,000-cell scale before separately increasing dataset size) to
disentangle whether structural-diversity count or dataset scale drives
the v0.3.1 collapse, rather than increasing both at once as done here,
deferring further 3D Swin Voxel Encoder fine-tuning-strategy refinements
(e.g., layer-wise learning rates, longer warmup) until that curriculum is
in place; (iii) investigate whether a lower InfoNCE temperature, a
mitotic-focused auxiliary loss, or a larger mitotic-oversampling fraction
can push anomaly-scoring AUC past the 0.70 threshold established in
v0.1.0/v0.2.0; (iv) extend to a three-tower
Voxel$\times$Text$\times$Gene/Structure-Identity contrastive objective
(``Medical CLAIP'', v0.4.0); (v) replace template-generated finding
text with LLM-expanded or, eventually, real pavement-inspection report
text; (vi) add a retrieval-augmented free-text generation downstream task
atop the diagnosis-reason retrieval introduced here; and (vii) begin
collecting and annotating real road-pavement 3D microscope images once
this methodology is fully validated.

All code and configuration files for this MVP are available at
\url{https://github.com/tk-yasuno/cell_3dswin_cl}.

\section{Supplementary Material: Extension to Structural Diversity and
         Dataset Scale (v0.3.1)}
\label{sec:v031}

We report this secondary v0.3.1 experiment as supplementary material,
rather than in the main text, because its outcome is a negative
generalization-limit result (Section~\ref{sec:v031collapse}) rather than
a core positive contribution of this MVP.

Having established in v0.3.0 that full fine-tuning of the Voxel Encoder
resolves the LoRA-on-random-backbone collapse (Section~\ref{sec:lessons2}),
we next ask whether that same recipe is \emph{robust} as structural
diversity and dataset scale increase -- a natural generalization question
for any methodology intended to eventually span many pavement-material
damage categories rather than a handful of proxy classes.

\subsection{Setup}

We extend the corpus from 4 to 8 structures, adding FBL (nucleolus),
SEC61B (endoplasmic reticulum), ST6GAL1 (Golgi apparatus), and ACTB (actin
filaments) to v0.3.0's LMNB1/TOMM20/TUBA1B/TJP1, chosen to maximize
organelle-category diversity (nuclear substructure, secretory pathway, and
a second, distinct cytoskeletal network alongside microtubules). We also
increase \texttt{per\_class\_target} from 750 to 1{,}000 cells/structure
(8{,}000 cells total: 6{,}400 train+val pool + 1{,}600 fixed test, 200
cells/structure test split, up from 125 in v0.3.0), specifically to
secure more mitotic
(abnormal-like) samples for anomaly-scoring validation (realized
abnormal-like count: 334/1{,}600 in the test split, up from 114/500 in
v0.3.0). The architecture, LoRA/full-fine-tune policy, optimizer, and
per-epoch budget (15 epochs) are otherwise unchanged from v0.3.0
(Section~\ref{sec:contrastive}). Progressive scales are redefined over the
larger 6{,}400-cell pool as 1.6k/3.2k/4.8k/6.4k (25/50/75/100\%); due to
compute/time constraints, training and evaluation stop at 4.8k and 6.4k is
not run.

\subsection{Results}
\label{sec:v031results}

\begin{figure}[h]
\centering
\includegraphics[width=0.47\textwidth]{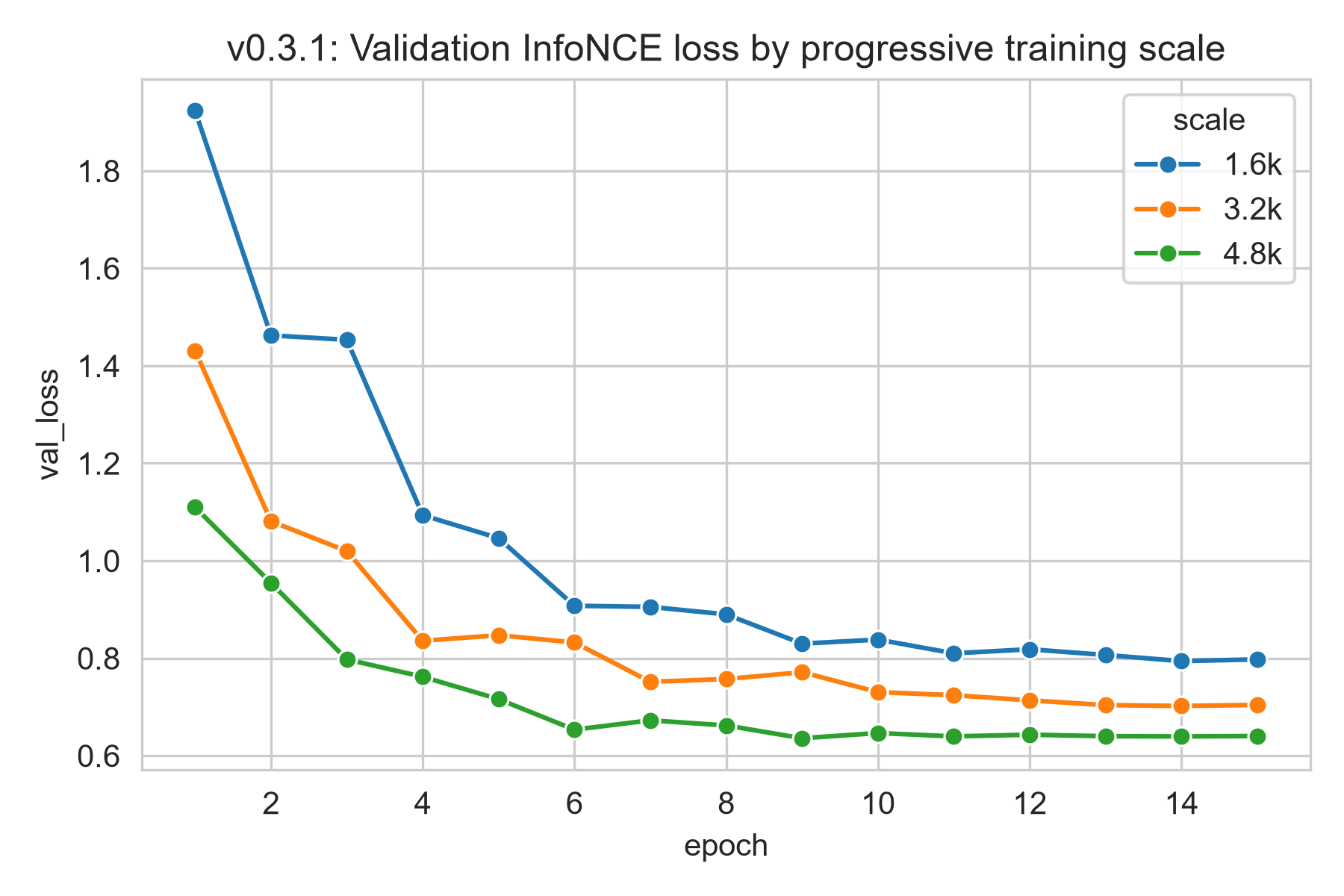}
\caption{v0.3.1: validation InfoNCE loss by progressive training scale.
Unlike the collapsed downstream metrics below, the loss curves are
smoothly and monotonically ordered by scale (1.6k $>$ 3.2k $>$ 4.8k at
essentially every epoch), the clearest piece of evidence that the
training/optimization pipeline scales as expected even as the learned
voxel representation fails to become structure-discriminative.}
\label{fig:v031valloss}
\end{figure}

\begin{figure}[h]
\centering
\includegraphics[width=0.47\textwidth]{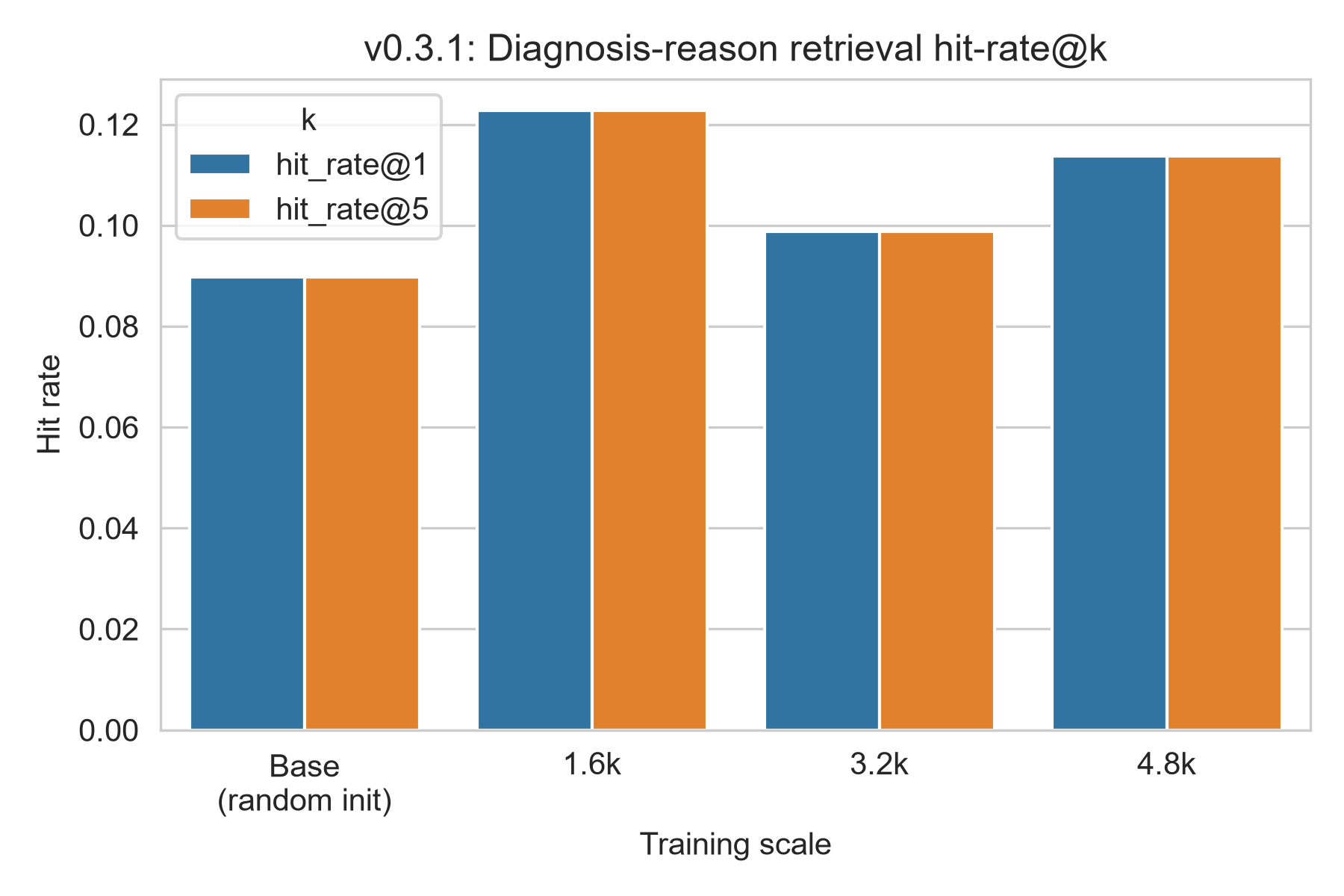}
\caption{v0.3.1: diagnosis-reason retrieval hit-rate@\{1,5\} by scale.
Unlike zero-shot accuracy's exact chance-level plateau, hit-rate exceeds
the random-init base (9.0\%) at every trained scale (12.3\%/9.9\%/11.4\%
at 1.6k/3.2k/4.8k), a modest but consistent positive signal suggesting
the joint structure+mitotic-state retrieval task retains some residual
learnable signal even under voxel-embedding collapse.}
\label{fig:v031diaghit}
\end{figure}

Table~\ref{tab:v031results} summarizes all four downstream evaluations at
each scale. Training loss decreases monotonically and substantially with
scale (val.\ loss 0.798$\to$0.704$\to$0.640 at 1.6k/3.2k/4.8k, each lower
than the corresponding v0.3.0 value at a similar relative pool fraction),
with the three per-epoch curves staying cleanly and monotonically ordered
by scale throughout training (Figure~\ref{fig:v031valloss}) -- confirming
the training/optimization pipeline itself scales correctly even though,
as detailed below, the resulting representation does not. Yet 8-way
zero-shot structure accuracy is \emph{exactly} at chance
(12.5\% $=$ 200/1{,}600) at every trained scale, and bidirectional
retrieval recall remains near the base level (R@10 $\le$0.9\%) throughout.
Voxel anomaly-scoring ROC-AUC improves mildly with scale (0.518
$\to$0.491$\to$0.514$\to$0.604 at base/1.6k/3.2k/4.8k) but, as in v0.3.0,
does not clear the 0.70 threshold. Diagnosis-reason retrieval hit-rates
are low and non-monotonic (9.0\%/12.3\%/9.9\%/11.4\%), but, unlike
zero-shot accuracy, every trained scale exceeds the random-init base
(Figure~\ref{fig:v031diaghit}) -- a second partial-progress signal amid
an otherwise negative result.

\begin{table}[h]
\centering
\caption{v0.3.1 (8 structures, 8,000 cells) downstream results by scale.}
\label{tab:v031results}
\footnotesize
\setlength{\tabcolsep}{3pt}
\begin{tabular}{lrrrr}
\toprule
\textbf{Scale} & \textbf{ZS acc.} & \textbf{R@10} & \textbf{Anom.\ AUC} & \textbf{Diag.\ hit@1} \\
\midrule
Base & 7.6\%  & 0.7\% & 0.518 & 9.0\%  \\
1.6k & 12.5\% & 0.9\% & 0.491 & 12.3\% \\
3.2k & 12.5\% & 0.7\% & 0.514 & 9.9\%  \\
4.8k & 12.5\% & 0.5\% & 0.604 & 11.4\% \\
\bottomrule
\end{tabular}
\end{table}

\begin{figure}[h]
\centering
\includegraphics[width=0.47\textwidth]{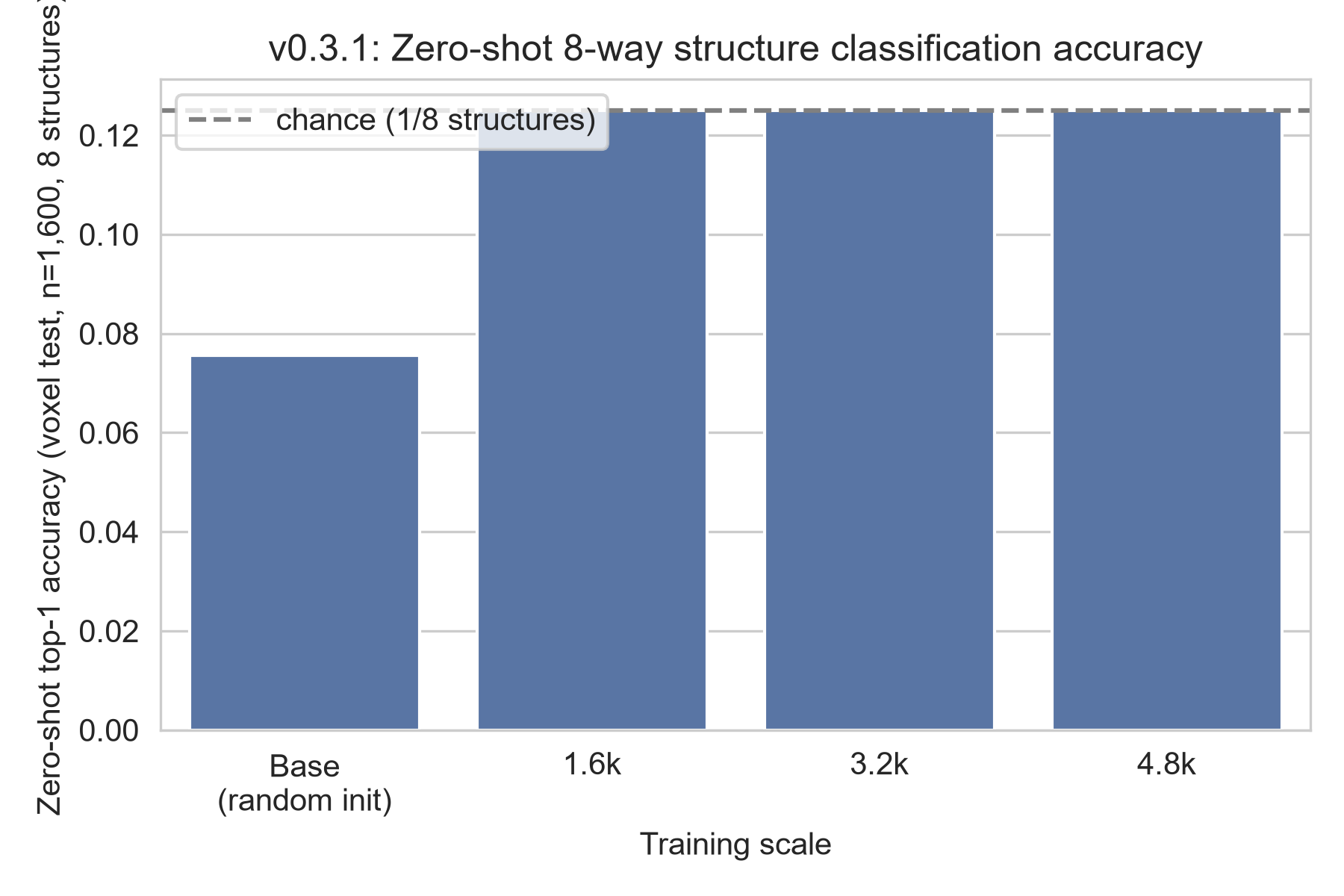}
\caption{v0.3.1: 8-way zero-shot structure classification accuracy by
scale, pinned exactly at the 12.5\% chance line for every trained scale
(cf.\ Figure~\ref{fig:zeroshot}'s clear v0.3.0 improvement).}
\label{fig:v031zeroshot}
\end{figure}

\begin{figure}[h]
\centering
\includegraphics[width=0.47\textwidth]{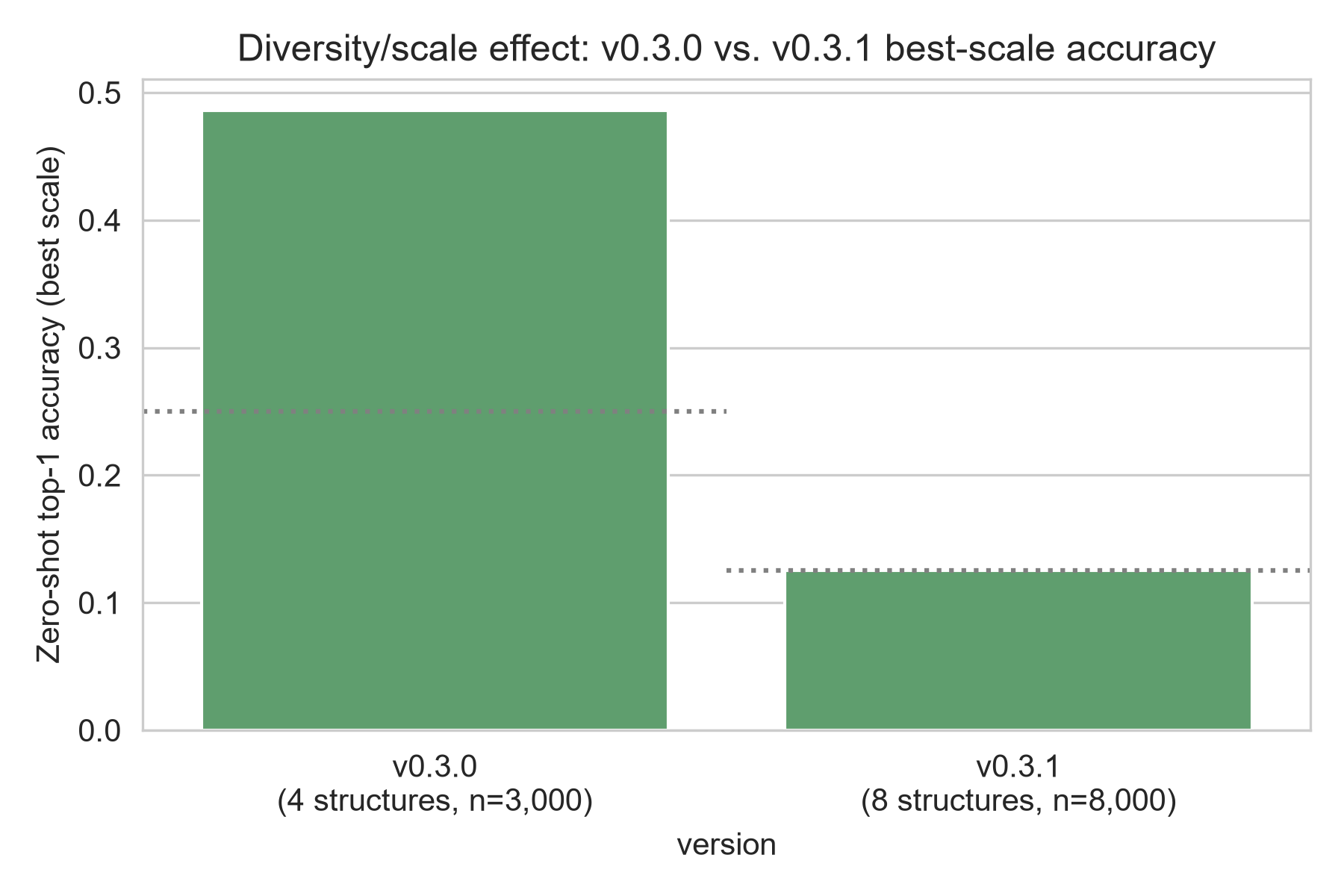}
\caption{Direct before/after comparison: v0.3.0's best scale (2.5k, 4
structures) reaches 48.6\% zero-shot accuracy, while v0.3.1's best
trained scale (4.8k, 8 structures) remains at exact chance (12.5\%).}
\label{fig:v03vsv031}
\end{figure}

\subsection{The Collapse Recurs: A Generalization-Limit Finding}
\label{sec:v031collapse}

Zero-shot accuracy pinned exactly at $1/8$ for every trained scale is the
same diagnostic signature as v0.3.0's original LoRA-on-random-backbone
failure (Section~\ref{sec:lessons2}): a single constant prediction for
every test example. Repeating the embedding-collapse probe (pairwise
voxel-embedding cosine similarity across distinct structures) on the 4.8k
checkpoint confirms this directly -- similarities of 0.92--0.998 across
\emph{all} structure pairs, versus the clear same-structure
(0.98--0.997) vs.\ cross-structure (0.79--0.93) separation observed in
v0.3.0's successful 2.5k checkpoint (Section~\ref{sec:lessons2}). Critically,
the Voxel Encoder in this run uses the \emph{identical} full-fine-tuning
policy that resolved the collapse in v0.3.0 -- no architectural regression
occurred. Two observations help localize the cause. First, the
corresponding ClinicalBERT text embeddings remain well-separated by
structure (pairwise cosine similarity $-0.46$ to $0.36$ across the 8
display-name prompts), so the Finding Encoder continues to function
correctly; the collapse is specific to the Voxel Encoder. Second, training
loss fell to 0.640 -- markedly \emph{lower} than v0.3.0's comparable-scale
values -- despite the voxel tower producing near-input-invariant
embeddings. This indicates the InfoNCE objective was satisfied
predominantly through a shortcut rather than through learning the
intended structure-identity signal, most plausibly the much stronger,
template-driven interphase/mitotic wording difference in finding text,
with which anomaly-scoring's mild AUC improvement at 4.8k is also
consistent. We did not
have compute/time budget remaining (per the scope decision in
Section~\ref{sec:v031}) to run the ablations needed to confirm this
hypothesis conclusively (e.g., per-epoch embedding-diversity probes,
learning-rate or batch-size sweeps, or an auxiliary structure-classification
loss), so we report it as an open negative result rather than a resolved
one.

\subsection{Implications for Generalization}

This result tempers v0.3.0's conclusion that full fine-tuning of a
randomly initialized 3D Swin voxel backbone is a fully general fix. It
is a necessary correction relative to LoRA-only adaptation (which failed
unconditionally at 4 structures), but it is not, on this evidence,
\emph{sufficient} to guarantee convergence to the intended structure-
discriminative representation as class count and dataset scale increase:
with 2$\times$ the structures and 2.67$\times$ the cells, the same
15-epoch training budget converged to a different, lower-loss but less
useful optimum at every scale tested. We regard this as the central
diversity/generalization finding of this MVP extension: \textbf{the
contrastive-learning methodology's reliability is scale- and diversity-
dependent, and verifying downstream task performance (not just training
loss) at each new scale/diversity regime is not optional} -- a stronger
version of the same lesson from Section~\ref{sec:lessons2}, now shown to
recur even after applying its own fix.

In hindsight, we also regard the \emph{design} of this extension, not
only its outcome, as part of the lesson. v0.3.1 increased structural
diversity (4$\to$8 structures) and dataset scale (3,000$\to$8,000 cells)
simultaneously, in a single step, rather than staging either change
incrementally (e.g., first validating 6 structures at the original
3,000-cell scale, then 8 structures at the same scale, before separately
scaling up dataset size). This confound means the evidence collected here
cannot distinguish whether structural-diversity count, dataset scale, or
their interaction is responsible for the collapse recurring
(Section~\ref{sec:v031collapse}). Given this, and the compute/time budget
already spent on this MVP extension, we treat a properly staged
diversity/scale curriculum -- together with the per-epoch
embedding-diversity probes and other ablations needed to confirm
Section~\ref{sec:v031collapse}'s shortcut-learning hypothesis -- as
future work, and explicitly defer further 3D Swin Voxel Encoder
fine-tuning-strategy refinements (e.g., layer-wise learning rates, longer
warmup, or an auxiliary structure-classification loss) beyond this MVP's
current scope. As noted in Section~\ref{sec:v031results}, the training
pipeline and the joint structure+mitotic-state retrieval task both retain
some partial, exploitable signal despite the collapse, which a future,
more carefully staged run could build on.

\bibliographystyle{unsrt}
\bibliography{methodology}

\end{document}